\pdfoutput=1
\documentclass[11pt]{article}

\usepackage[final]{acl}

\usepackage{times}
\usepackage{latexsym}

\usepackage[T1]{fontenc}
\usepackage[utf8]{inputenc}

\usepackage{microtype}

\usepackage{inconsolata}

\usepackage{graphicx}
\usepackage{subfigure}
\usepackage{booktabs} % for professional tables
\usepackage{outlines}
\usepackage{subcaption}
\usepackage{wrapfig}

\definecolor{lightblue}{RGB}{200, 230, 255} % Soft pastel blue
\definecolor{lightred}{RGB}{255, 210, 210}

\usepackage{adjustbox}
\usepackage{booktabs}

\usepackage{hyperref}

\usepackage{amsmath}
\usepackage{amssymb}
\usepackage{mathtools}
\usepackage{amsthm}
\usepackage{multirow}
\usepackage{makecell}
\usepackage{tikz}
\usepackage{adjustbox}

\usepackage[capitalize,noabbrev]{cleveref}

\theoremstyle{plain}

\theoremstyle{definition}

\theoremstyle{remark}

\usepackage[textsize=tiny]{todonotes}

\title{Pixels Versus Priors: Controlling Knowledge Priors in Vision-Language Models through Visual Counterfacts}

\author{
    Michal Golovanevsky$^{\star 1}$, William Rudman$^{\star 1}$, Michael Lepori $^{1}$, Amir Bar$^{2}$, \\
    {\bf Ritambhara Singh$^{1}$},
    {\bf Carsten Eickhoff$^{3}$}  \\
    $^{1}$Brown University, $^{2}$Tel Aviv University, $^{3}$University of Tübingen \\
    \texttt{\{michal\_golovanevsky, william\_rudman\}@brown.edu} \\
}

\begin{document}
\maketitle
\begin{abstract}
\renewcommand{\thefootnote}{\fnsymbol{footnote}}
 
Multimodal Large Language Models (MLLMs) perform well on tasks such as visual question answering, but it remains unclear whether their reasoning relies more on memorized world knowledge or on the visual information present in the input image. To investigate this, we introduce Visual CounterFact, a new dataset of visually-realistic counterfactuals that put world knowledge priors (e.g, red strawberry) into direct conflict with visual input (e.g, blue strawberry). Using Visual CounterFact, we show that model predictions initially reflect memorized priors, but shift toward visual evidence in mid-to-late layers. This dynamic reveals a competition between the two modalities, with visual input ultimately overriding priors during evaluation. To control this behavior, we propose \textit{Pixels Versus Priors} (PvP) steering vectors, a mechanism for controlling model outputs toward either world knowledge or visual input through activation-level interventions. On average, PvP successfully shifts 99.3\% of color and 80.8\% of size predictions from priors to counterfactuals. Together, these findings offer new tools for interpreting and controlling factual behavior in multimodal models. 
\footnote{Equal contribution. Order determined by coin flip.} 
\footnote{Code: \href{https://github.com/rsinghlab/pixels_vs_priors}{
\textit{https://github.com/rsinghlab/pixels\_vs\_priors}}}

\end{abstract}

\section{Introduction}

\begin{figure}[h]
    \centering
    \includegraphics[width=\columnwidth]{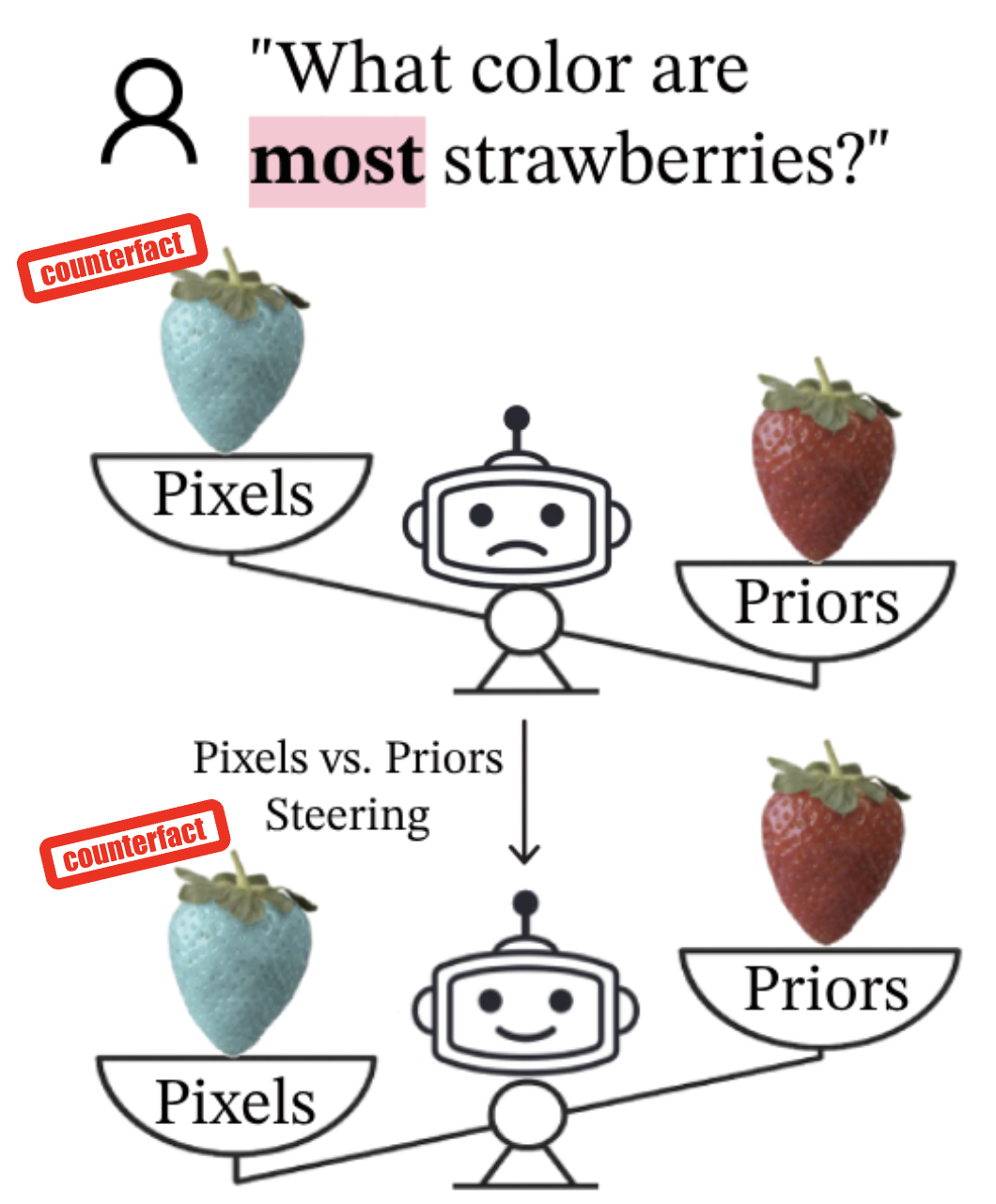}
    \caption{\textbf{Pixels Versus Priors Steering.} We introduce a framework for controlling whether a vision-language model relies on visual input or memorized knowledge. Counterfactual visual evidence often overrides world knowledge priors.}
    \label{fig:pixels_priors}
\end{figure}

As multimodal large language models (MLLMs) demonstrate increasing success in real-world vision-language tasks \cite{llava-next, qwen2-vl, janus-pro}, it is becoming increasingly important to understand their internal mechanisms in order to ensure the reliability and safety of these systems \cite{notice, jiang2024interpreting, luo2024task, rudman2025forgotten}. Despite recent advances, interpretability in MLLMs remains underdeveloped compared to progress in natural language processing (NLP), where the encoding of world knowledge facts is well-researched and methods exist for systematically editing factual associations \cite{rome}.
In NLP, \textit{counterfactual datasets} consist of minimally altered input pairs that isolate specific factual changes, such as swapping one entity or relation while holding others constant. These datasets enable causal analysis of model behavior and have been central to understanding how factual associations are stored, retrieved, and manipulated \cite{geva2020transformer, geva2023dissecting, dai2021knowledge, yu2023characterizing, rome}. Unlike language, where factual associations are well understood, there is no visual equivalent for locating or modifying stored associations in MLLMs. In particular, there is no counterfactual dataset for testing how these models balance visual perception against memorized priors, nor any method for controlling their responses when the two sources of information conflict. To address this gap, we introduce Visual CounterFact, the first dataset designed to study world knowledge priors related to visual attributes in MLLMs, and use it to develop \textit{Pixels Versus Priors} steering (PvP), a method for controlling whether the model relies on pixel-level information or on world knowledge.

\textit{Visual CounterFact} modifies visual attributes, color and size, of everyday objects to create direct conflicts between memorized facts and input pixels. In our framework, \textit{world knowledge priors} refer to linguistic associations between visual attributes and objects that the model has memorized during pretraining. In contrast, \textit{visual perception} is defined by the in-context visual input, which we manipulate to create counterfactual images. These counterfactuals are designed to challenge the model’s world knowledge of visual attributes by presenting plausible but contradictory visual evidence. For example, we contrast the size-related knowledge prior ``strawberries are bigger than flies'' with the counterfactual ``flies are bigger than strawberries,'' violating expected size relations (see Figure~\ref{fig:counterfactuals}).

Using Visual CounterFact, we find that MLLMs often ignore world knowledge when shown counterfactual images, favoring perceptual input even when prompted for general facts. We then trace where in the forward pass predictions shift from in-weight knowledge (e.g., strawberries are red) to in-context perception (\textit{this} strawberry is blue), finding that this transition consistently emerges in mid-to-late layers. During this transition, models frequently flip between the two answers, revealing a competition between in-context pixel and in-weights prior information, with pixels often overriding priors in the model's output. To control this behavior, we use our \textit{Pixels Versus Priors} steering to control whether a vision-language model relies on knowledge priors or pixel information. PvP is a novel framework to construct steering vectors for vision-language models that control whether the model responds based on memorized knowledge or in-context visual input. Through this steering, we successfully shift an average of 99.3\% of color predictions and 80.8\% of size predictions from memorized priors to counterfactual answers.
%Notably, steering from counterfactual to world knowledge is significantly more challenging, suggesting that visual perception tends to override knowledge priors.

Together, these contributions provide a new visual counterfactual benchmark and a mechanism for interpreting and controlling the behavior of vision-language models. We present the necessary foundation for a mechanistic understanding of how MLLMs integrate image input with prior knowledge of visual attributes, bridging the gap between interpretability research in language models and the emerging needs of multimodal models.

% \begin{figure*}[h]
%     \centering
%     \includegraphics[width=0.8\textwidth]{figures/task_vector_diagram_2.png}
% \caption{Pixel versus Prior create steering vectors by contrasting representations of prompts that emphasize pixel-level information (``this'') versus priors (``most''). We use the last hidden state compute the steering vectors.}
%     \label{fig:task_vectors}
% \end{figure*}

\section{Related Works}

Studies in mechanistic interpretability have shown that LLMs encode factual associations grounded in world knowledge within their weights, enabling precise manipulation through targeted interventions.
In particular, feedforward layers often act as key-value memories, injecting factual knowledge into subject representations \cite{geva2020transformer, geva2023dissecting}, while clusters of ``knowledge neurons'' have been shown to store and control specific facts \cite{dai2021knowledge, yu2023characterizing}. These internal representations can be edited by introducing counterfactuals through weight-level interventions\cite{rome}, or by tracing how attention mechanisms recover or suppress modified content during inference \cite{jin2024cutting}. More recently, activation-level interventions have emerged as an alternative to weight-level editing. In NLP, \textit{steering vectors} are computed by subtracting internal representations from contrasting prompts to isolate meaningful activation directions \cite{subramani2022extracting, turner2023steering}. These directions can be added at inference to shift a model’s behavior without altering the model’s weights. 

Although model editing and steering have been successful in shifting model outputs, \citet{gekhman2025inside} show that even when a model produces incorrect outputs, it may still internally represent the correct fact, highlighting a disconnect between stored knowledge and in-context generation. Investigating this disconnect further, there is a growing body of NLP literature that seeks to understand how language models flexibly deploy both in-context and memorized in-weight knowledge \cite{chan2022data, singh2023transient, anand2024dual, reddymechanistic, lampinen2024broader, zucchet2025language, park2024competition}. These studies suggest that models often switch between relying on memory and adapting to context, depending on training dynamics and task structure. This inherent conflict \textit{within}-modality motivates our mechanistic analysis \textit{across}-modalities.

In order to study the conflict of world knowledge priors across vision and language, a visual counterfactual dataset is needed. While many benchmarks test visual-textual alignment, none directly evaluate a model's reliance on visual world knowledge. VL-Checklist \cite{zhao2022vl} and VALSE \cite{parcalabescu2021valse} vary captions over fixed images, while FOIL-COCO \cite{shekhar2017foil}, SVO-Probes \cite{hendricks2021probing}, and Winoground \cite{thrush2022winoground} alter semantic content in real images. However, these methods often suffer from uncontrolled visual artifacts. COCO-Counterfactuals \cite{le2023coco} uses generative models to edit images to replace a single object in the image with a new object. For example, for an image reflecting the caption ``A large black ball sitting next to a glass of \textbf{milk}'', they generate a ``counterfactual'' image from the prompt ``A large black ball sitting next to a glass of \textbf{water}''. While these images represent minimally altered pairs, they are not true counterfactuals designed to contradict visual world knowledge priors, such as the expectation that ``strawberries are red'' or that ``strawberries are larger than flies''.

Similarly, model editing methods developed for LLMs have proven difficult to adapt to vision-language models. Early work shows that multimodal neurons can encode visual-textual concepts \cite{pan2023finding}, and that vision and language encoders often share object-level semantics \cite{sammani2024interpreting}. Yet, attempts to localize or edit factual knowledge in MLLMs, such as MMEdit \cite{cheng2023can} and VLKEB \cite{huang2024vlkeb} face challenges with generalization and control. Model edits can introduce unintended side effects, such as altering predictions on unrelated inputs, and often fail to generalize across paraphrased prompts or unseen contexts. In addition to the side effects of model editing, in this work, we show that MLLMs tend to override memorized knowledge priors when presented with conflicting visual input. Given that pixels override priors, applying model editing to steer the model towards ``strawberries are blue'' would be overridden when presented with an image of a red strawberry. Instead, steering provides a reliable mechanism to modulate model responses between pixels and priors.

Steering has seen little adoption in the multimodal setting. \citet{steering_hallucinations} propose latent-space steering to reduce hallucinations by stabilizing vision features at inference time. \citet{soft_prompts} show that embedded image prompts can act as hidden meta-instructions, influencing output style or sentiment. \citet{luo2024task} demonstrate that vision-language models learn shared task vectors that generalize across modalities, suggesting that internal representations can be steered with compact task encodings. Similarly, \citet{hojel2024finding} identify visual task vectors in the activation space of prompting models, showing that these representations can be patched into attention heads to guide model behavior across tasks. However, these works do not examine how models handle conflicts between visual input and stored knowledge, nor do they provide mechanisms for explicitly controlling which source of information the model relies on. To our knowledge, no prior work applies steering vectors to vision-language models for the purpose of modulating reliance on memorized visual priors versus image inputs.

\section{Creating Visual CounterFact}
\begin{figure}[h]
    \centering
    \includegraphics[width=\columnwidth]{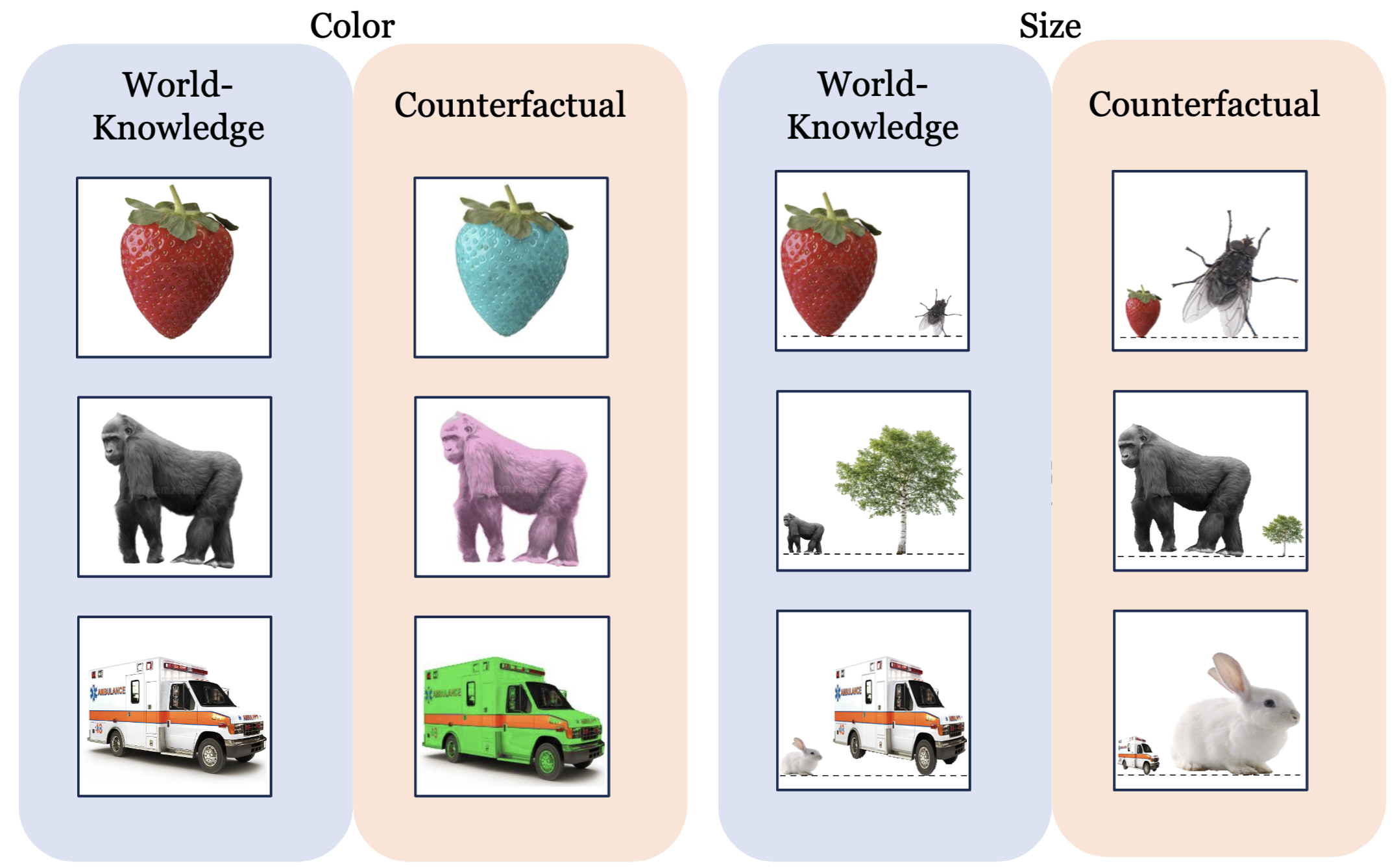}
    \caption{\textbf{Visual CounterFact.} A new benchmark to study how VLMs utilize world knowledge compared to visual inputs. (Left) images created using color relations, 
    (right) images created using size relations.}
    \label{fig:counterfactuals}
\end{figure}

First, we describe the creation of \textit{Visual CounterFact}, a dataset designed to examine how MLLMs use visual input and world knowledge when presented with controlled counterfactual examples. \textit{Visual CounterFact} contains images that deliberately introduce conflicts between visual input and world knowledge, spanning two tasks: color and size. Each image is created through a four-step pipeline designed to preserve realism and control for background noise while introducing counterfactual evidence. Additional details on each step are provided in Appendix Section \ref{app:dataset_construction}.

\textbf{(Step 1) Identifying objects with strong visual priors.} We begin by selecting objects that have widely known visual attributes, such as canonical colors (e.g., ``strawberries are red'') or typical size relationships (e.g., ``strawberries are larger than flies''). These objects are sourced from human-annotated datasets (McRae norms \cite{mcrae2005semantic}) and extended with GPT-4o estimates of typical attributes for CIFAR-100 \cite{cifar100}  and ImageNet \cite{imagenet} categories.

\textbf{(Step 2) Retrieving world knowledge images.} For each object, we collect images using the Google Images API, specifying that the object should appear on a white background to reduce spurious visual cues. We aim to retrieve images that match the canonical visual prior (e.g., a red strawberry rather than a pink or green one). Each image is filtered and scored by GPT-4o for color accuracy, object correctness, and realism, and the highest-scoring image is selected.

\textbf{(Step 3) Generating counterfactual relations.} We construct counterfactuals that intentionally conflict with typical visual priors for each object. For the color task, we first prompt the LLaVA-Next model to generate likely colors for a given object (e.g., “What color is a strawberry?”), then sample from the five least likely common colors (e.g., blue, orange, purple) to select a counterfactual color. To maintain visual clarity, we constrain these counterfactuals to be visually distinct from the original (e.g., avoiding red/orange or gray/black swaps). For the size task, we use GPT-4o to estimate the real-world dimensions of objects and compute their total size. We select object pairs that differ by at least a factor of 10 and generate two counterfactual relations per object by inverting the expected size ordering. For example, if object A is smaller than object B and object B is smaller than object C, we create counterfactuals such as ``A is bigger than B'' and ``B is bigger than C.'' 

\textbf{(4) Editing images to reflect counterfactual attributes.} We use SAM2 \cite{sam2} segmentation masks to apply controlled, localized transformations. In the color task, we modify hue values while preserving texture and shading (Figure \ref{app:dataset_examples} Color); in the size task, we resize object masks and align them on a dashed line to reflect altered size relations without introducing depth ambiguity (Figure \ref{app:dataset_examples} Size). The final dataset contains 575 color exemplars, 575 color counterfactuals, and 877 original and 877 counterfactual size images, for a total of 2,904 visually grounded examples. 

\section{Methods}

We use Visual CounterFact to evaluate how MLLMs store world knowledge priors of visual characteristics using three models: LLaVA-Next-7B \cite{llava-next}, Qwen2-VL-7B \cite{qwen2-vl}, and DeepSeek Janus Pro-7B \cite{janus-pro}. These models were selected to cover a range of current state-of-the-art multimodal architectures, including established families like LLaVA, and emerging MLLMs like QwenVL and Janus Pro. We first apply early decoding to trace the evolution of the model’s prediction across layers and identify the point at which visual information overtakes linguistic priors or vice versa. We then develop PvP steering vectors that can actively shift model behavior toward either image-grounded or world knowledge responses. The results of these methods are presented in Sections~\ref{sec:early_decoding_results} and~\ref{sec:steering_results}, respectively.

\subsection{Early Decoding}

Early decoding is a technique for probing the intermediate computations of a model by decoding hidden states before the final output layer. Originally introduced by \citet{nostalgebraist2020logitlens} and extended in follow-up work \citep{belrose2023eliciting, Pal_2023, ghandeharioun2024patchscopes, vilas2023analyzing}, this method applies the final layernorm, $\sigma$, to the intermediate hidden states $h_l$ at layer $l$ and then projects this representation onto the vocabulary space using the unembedding matrix $W_U$, yielding $W_U( \sigma(h_l))$. This produces a probability distribution over tokens, effectively allowing us to observe what the model ``believes'' at a given stage in its forward pass.

We use early decoding to identify when the model’s prediction shifts from being guided by knowledge stored in weights to being grounded in visual perception. By decoding the model’s predictions layer by layer, we observe how the probability distribution over possible output tokens evolves, allowing us to pinpoint where the model begins to favor a counterfactual (image-based) answer over the memorized world knowledge alternative.

%At each layer $l$, we extract the hidden state $h_l$ and measure the probability of both the world knowledge and the counterfactual answer by applying the projection $W_U(\sigma(h_l))$, where $\sigma$ denotes the final layer-norm. Since each model considered in this study uses residual connections that accumulate representations across layers, early decoding provides a continuous trace of how the model's internal predictions evolve as it integrates textual and visual information.

\subsection{Pixel Versus Prior Steering}
\label{sec:pvp_steering}
\begin{figure*}[h]
    \centering
    \includegraphics[width=\textwidth]{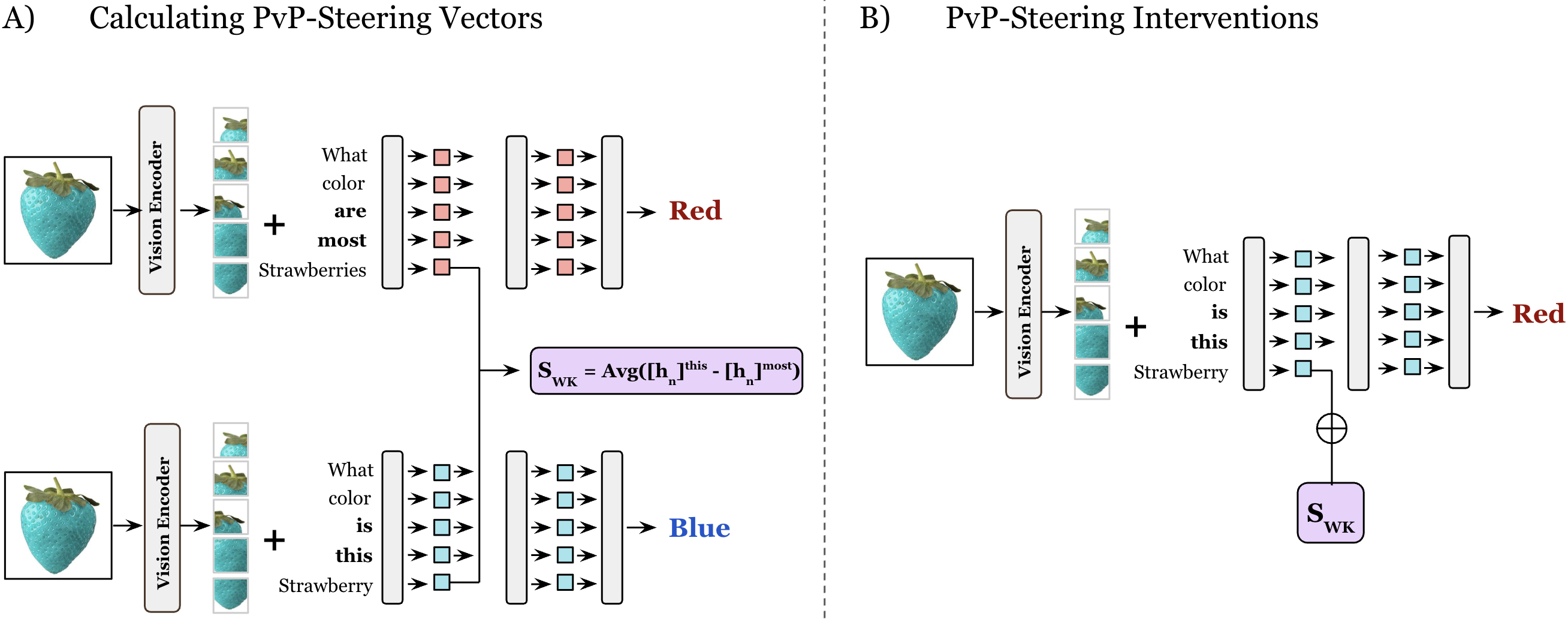}
\caption{Pixel Versus Prior steering vectors are created by contrasting representations of prompts that emphasize pixel-level information (``this'') versus priors (``most''), using the last hidden state.}
    \label{fig:task_vectors}
\end{figure*}
Using Visual CounterFact, we introduce \textit{Pixels Versus Priors} (PvP) steering vectors by calculating the difference in activations with contrasting prompts. Specifically, we present the model with a counterfactual image accompanied by one prompt that encourages the retrieval of world knowledge priors and another that directs it to analyze the image pixels. Consider the example in Figure~\ref{fig:task_vectors}. The prompt ``What color \textit{is this} strawberry?'' encourages a visually grounded response, while ``What color \textit{are most} strawberries?'' draws on memorized world knowledge priors about the color of a strawberry. When paired with a counterfactual image (e.g., a blue strawberry), the model should ideally answer ``blue'' in the first case and ``red'' in the second. When computing PvP steering vectors, the visual input is always the \textit{counterfactual image}. For a given layer, $l$, we extract the hidden representations at the MLP block for both prompts at each layer and compute two steering vectors, $S^{l}_{CF}$ and $S^{l}_{WK}$:
\[
S^{l}_{CF} = \frac{1}{D}\sum_{i}^{D} ([h_{n}^{l}]^{\text{this}}_{i} - [h_{n}^{l}]^{\text{most}}_{i}) \]
\[S^{l}_{WK} =  \frac{1}{D}\sum_{i}^{D}([h_{n}^{l}]^{\text{most}}_{i} - [h_{n}^{l}]^{\text{this}}_{i}).
\]
Here, $i \in \{1,2...,D\}$ represent the text-image pairs in Visual CounterFact and $h_{n}$ represents the hidden state of the last text token in the sequence of a sample, typically an instruction token, which has been shown to store more important information when compared to specific subject tokens \cite{notice}. After computing the world knowledge ($S^{l}_{\text{WK}}$) and counterfactual ($S^{l}_{\text{CF}}$) steering vectors, we control the model’s output by modifying the hidden state of the final token in the sequence at a given layer in the language decoder. Formally, let $h_{n}^{l}$ denote the hidden state of the last token at layer $l$ in the language decoder of an MLLM. To steer the representation toward pixel-level information from the image, we apply the following intervention:
% \[
% [h_{n}^{l}]_{i}^{\text{most}} = [h_{n}^{l}]_{i}^{\text{most}} + S^{l}_{\text{CF}}
% \]
\[
\hat{h}_{n}^{l} = h_{n}^{l} + S^{l}_{\text{CF}}
\]
To instead steer the model toward world knowledge priors, we apply:
% \[
% [h_{n}^{l}]_{i}^{\text{this}} = [h_{n}^{l}]_{i}^{\text{this}} + S^{l}_{\text{WK}}
% \]
\[
\hat{h}_{n}^{l} = h_{n}^{l} + S^{l}_{\text{WK}}
\]
These interventions are applied for all $l \in [l, l+w]$. Our method for calculating multimodal steering vectors captures the representational shift needed to modulate the model’s reliance on vision input versus world knowledge priors.

\section{Results}

\subsection{MLLMs are Distracted By Counterfactual Images}\label{acc_results}

\begin{table*}[h]
\centering
%\scriptsize
\begin{tabular}{ll|rr|rr}
\toprule
\textbf{Model} & \textbf{Task} & 
\textbf{CF + ``this''} & \textbf{WK + ``this''} & 
\textbf{CF + ``most''} & \textbf{WK + ``most''} \\
\midrule
LLaVA-Next & Color & 85.19 & 87.22 & 47.26 & 92.09 \\
           & Size  & 82.12 & 96.42 & 40.30 & 95.60 \\
\midrule
Qwen2-VL   & Color & 84.79 & 85.40 & 60.65 & 90.87 \\
           & Size  & 91.20 & 98.21 & 28.34 & 96.29 \\
\midrule
Janus-Pro  & Color & 86.00 & 88.03 & 59.23 & 90.47 \\
           & Size  & 85.14 & 96.84 & 18.02 & 96.01 \\
\bottomrule
\end{tabular}

\caption{Accuracy (\%) for color and size tasks under ``this'' (e.g., ``What color is this strawberry?'') and ``most'' (e.g., ``What color are most strawberries?'') questions with counterfactual (CF) and world knowledge (WK) images. Models perform well when grounded in the current image, but accuracy drops sharply in the ``most + CF'' setting, indicating that MLLMs are overly influenced by misleading visual input.}
\label{tab:accuracies}
\end{table*}

We begin by analyzing how MLLMs behave when presented with counterfactual (CF) images that intentionally contradict common object priors, alongside baseline world knowledge (WK) images that reflect real-world visual expectations (Figure~\ref{fig:counterfactuals}). To test whether models rely more on memorized knowledge or on the current image, we use two types of prompts: ``What color are \textbf{most} <objects>?'' and ``What color is \textbf{this} <object>?''

All models perform well on ``this'' prompts, achieving over 80\% accuracy even when the input image presents a counterfactual. This indicates that MLLMs are highly effective at grounding their answers in the current visual input. Errors in this setting typically involve subtle hue disagreements such as gold versus orange or yellow, rather than confusion about the underlying object property.

In contrast, the ``most'' prompts reveal a critical weakness. When asked about what is generally true, models are expected to retrieve world knowledge rather than attend to the current image. This behavior holds when WK images are shown, but accuracy drops sharply when the same question is paired with CF images. In these cases, models often abandon their prior knowledge in favor of what is visually presented, even though the prompt clearly targets a generic concept. This suggests that MLLMs are easily distracted by the current image, even when instructed to generalize. 
%This observation motivates our core contribution: a mechanism to steer model behavior toward either memorized knowledge or perceptual evidence.

\subsection{Localizing Visual Perception Shifts through Early Decoding} \label{sec:early_decoding_results}
% \begin{table}[h]
% \centering
% \resizebox{\columnwidth}{!}{%
% \begin{tabular}{lccc}
% \toprule
% \multicolumn{4}{c}{\textbf{Size}} \\
% \midrule
%                  & LLaVA‑Next & Qwen2‑VL & Janus‑Pro \\
% \midrule
% \% Samples w/o Flip     & 43\%       & 77\%     & 66\%      \\
% \% Samples w/ Flip        & 57\%       & 23\%     & 34\%      \\
% Avg. $\#$ CF $\to$ WK         & 1.00       & 0.33     & 0.80      \\
% Avg. $\#$ WK $\to$ CF     & 1.23       & 0.87     & 0.92      \\
% \midrule
% \multicolumn{4}{c}{\textbf{Color}} \\
% \midrule
% \% Samples w/o Flip     & 42\%       & 75\%     & 89\%      \\
% \% Samples w/ Flip        & 58\%       & 25\%     & 11\%      \\
% Avg. $\#$ CF $\to$ WK       & 0.79       & 0.18     & 0.69      \\
% Avg. $\#$ WK $\to$ CF    & 1.24       & 0.88     & 0.50      \\
% \bottomrule
% \end{tabular}%
% }
% \caption{Flip statistics for size and color attributes with counterfactual images and prompts designed to elicit world knowledge responses. A ``flip'' occurs when the initially less probable response later exceeds the alternative.}
% \label{tab:early_decoding_table}
% \end{table}

\begin{table*}[h]
\centering
\resizebox{\textwidth}{!}{%
\begin{tabular}{l|ccc|ccc}
\toprule
 & \multicolumn{3}{c|}{\textbf{Size}} & \multicolumn{3}{c}{\textbf{Color}} \\
\cmidrule(lr){2-4} \cmidrule(lr){5-7}
 & LLaVA-Next & Qwen2-VL & Janus-Pro & LLaVA-Next & Qwen2-VL & Janus-Pro \\
\midrule
\% Samples w/o Flip & 45\% & 69\% & 70\% & 35\% & 71\% & 88\% \\
\% Samples w/ Flip  & 55\% & 31\% & 30\% & 65\% & 29\% & 12\% \\
Avg. $\#$ CF $\to$ WK & 1.02 & 0.47 & 0.76 & 0.84 & 0.18 & 0.56 \\
Avg. $\#$ WK $\to$ CF & 1.12 & 0.70 & 0.86 & 1.31 & 0.90 & 0.56 \\
\bottomrule
\end{tabular}%
}
\caption{Flip statistics for size and color attributes with counterfactual images and prompts designed to elicit world knowledge responses. A ``flip'' occurs when the initially less probable response later exceeds the alternative by at least 5\%.}
\label{tab:early_decoding_table}
\end{table*}

\begin{figure}[h]
    \centering
    \includegraphics[width=\columnwidth]{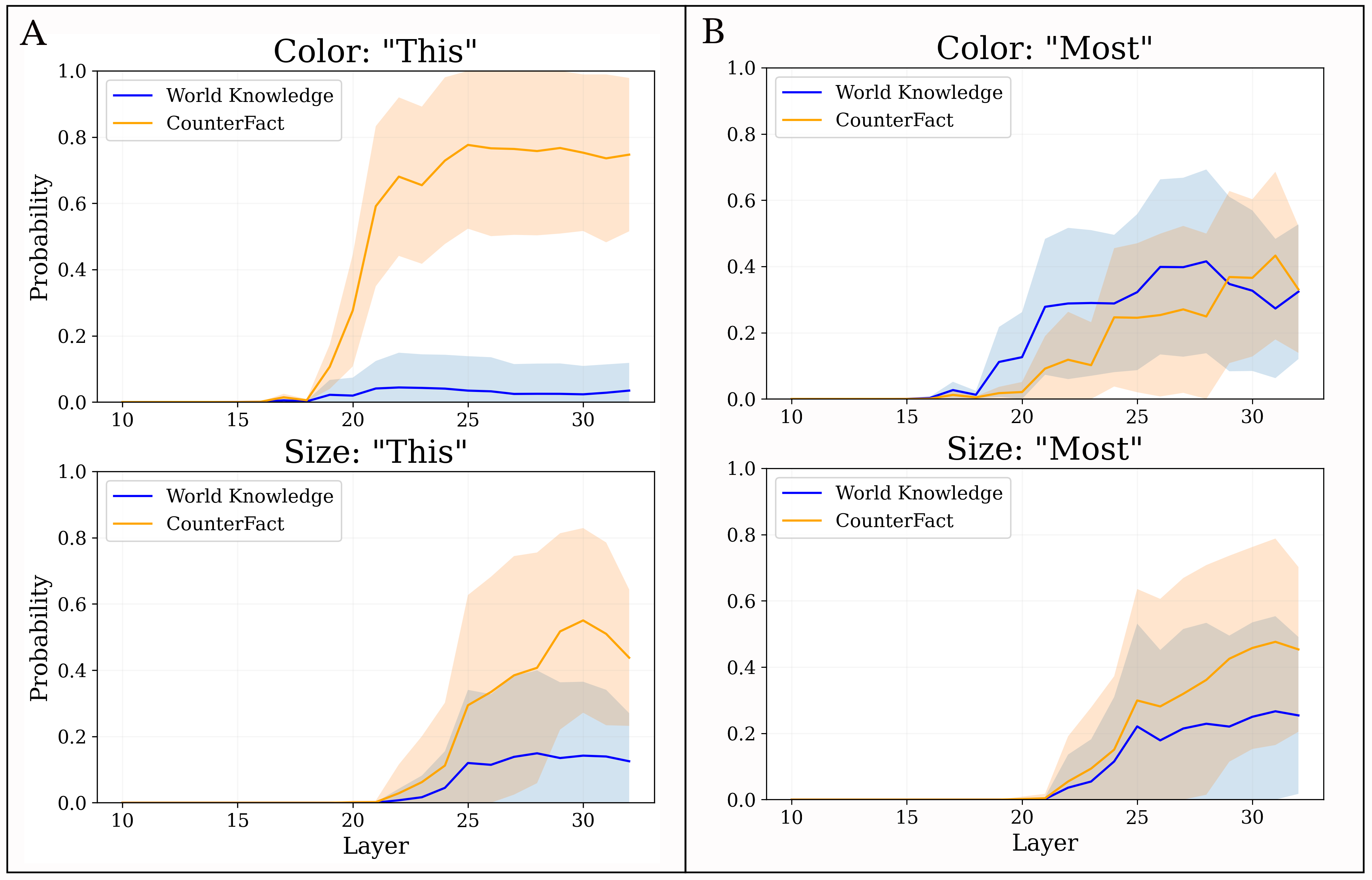}
    \caption{Early decoding results on LLaVA-Next show a conflict between answering ``world knowledge'' using priors or answering ``counterfact''.}
    \label{fig:early_decoding}
\end{figure}
 %Figure \ref{fig:early_decoding_appendix} results for QwenVL and Janus-Pro.
To understand how this visual override emerges during inference, we apply \textit{early decoding} to track model predictions across layers. This reveals when the model transitions from relying on memorized priors to integrating counterfactual visual input. Figure \ref{fig:early_decoding} shows that in the color task when the model is prompted for the world knowledge answer but given a counterfactual image, the probability of the world‐knowledge answer rises in mid to late layers, then flips to the counterfactual answer (orange) in the final layers. This ``flipping behavior'' is most common when the model is prompted to respond with the world knowledge answer and provided with a counterfactual image (Figure \ref{fig:early_decoding} Panel B). This delayed integration of visual input leads to errors when the image contradicts memorized associations, matching the results in Table \ref{tab:accuracies}. 

In contrast, when using ``this'' prompt (e.g., ``what color is \textbf{this} strawberry?'') for identifying the counterfactual attribute, models are confident in the counterfactual answer by the middle layers and rarely flip to the world‐knowledge alternative (Figure \ref{fig:early_decoding} panel A and Table~\ref{tab:early_decoding_table}). This confidence is supported by the high inference accuracies seen in Table \ref{tab:accuracies}. Despite their confidence in the counterfactual answer, there is still a slight spike in world knowledge answer probability in mid-to-later layers. This slight spike shows that memorized knowledge does not fully disappear from the model, even when presented with contradicting inputs. 

Table~\ref{tab:early_decoding_table} shows how often models alternate between world knowledge and counterfactual answers on the color and size tasks when provided with a counterfactual image and a ``most'' prompt. On average, LLaVA-Next flips from world knowledge to counterfact 1.24 times on samples where a flip occurs, compared to 0.79 in the reverse direction. This indicates that MLLMs are prone to overriding prior knowledge when presented with a counterfactual image. These results suggest a consistent pattern: models initially rely on linguistic priors rooted in world knowledge, and only later override these with visual evidence as processing progresses through the layers.

\subsection{Controlling World Knowledge Associations with PvP Steering Vectors}\label{sec:steering_results}

% \begin{table}[ht]
% \centering
% \begin{tabular}{l c c c c c}
% \hline
% \textbf{Task} & \multicolumn{2}{c}{\textbf{Color}} & \multicolumn{2}{c}{\textbf{Size}} \\
% \cline{2-5}
% \% flips      & WK $\to$ CF & CF $\to$ WK & WK $\to$ CF & CF $\to$ WK \\
% \hline
% \textbf{LLaVA-Next}         & 89.9   & 57.2 & 66.9 & 27.6 \\
% Key Layers & [16-19] & [11-18] & [10-18] & [14-23] \\
% \hline
% \textbf{QwenVL}             & 93.6   & 56.8 & 85.5 & 52.4 \\
% Key Layers & [18-21] & [17-22] & [17-24] & [11-22]\\
% \hline
% \textbf{Janus-Pro}          & 94.0   & 46.3 & 71.4 & 45.2 \\
% Key Layers & [17-20] & [16-19] & [15-18] & [17-21]\\
% \hline
% \end{tabular}
% \caption{Performance of models under Color and Size tasks with two flip directions: WK $\to$ CF and CF $\to$ WK. The key layers are shown in separate rows.}\label{tab:flips}
% \end{table}

\begin{table}[ht]
\centering
\small
\renewcommand{\arraystretch}{1.3} % Adjust this value to add more vertical space
\begin{tabular}{l l l c l}
\hline
\textbf{Model} & \textbf{Task} & \textbf{ Direction} & \textbf{Flips \%} & \textbf{Layers} \\
\hline
\multirow{4}{*}{\textbf{LLaVA}} 
  & \multirow{2}{*}{Color} & WK $\to$ CF & \textbf{99.5} & [14-16] \\
  &                       & CF $\to$ WK & 86.4 & [10-17] \\
  & \multirow{2}{*}{Size}  & WK $\to$ CF & \textbf{71.3} & [8-16] \\
  &                       & CF $\to$ WK & 33.5 & [12-21] \\
\hline
\multirow{4}{*}{\textbf{QwenVL}} 
  & \multirow{2}{*}{Color} & WK $\to$ CF & \textbf{99.7} & [17-19] \\
  &                       & CF $\to$ WK & 78.8 & [12-17] \\
  & \multirow{2}{*}{Size}  & WK $\to$ CF & \textbf{89.9} & [16-22] \\
  &                       & CF $\to$ WK & 61.8 & [13-23] \\
\hline
\multirow{4}{*}{\textbf{Janus-Pro}} 
  & \multirow{2}{*}{Color} & WK $\to$ CF & \textbf{98.6} & [14-16] \\
  &                       & CF $\to$ WK & 78.2 & [15-18] \\
  & \multirow{2}{*}{Size}  & WK $\to$ CF & \textbf{81.2} & [16-19] \\
  &                       & CF $\to$ WK & 70.37 & [16-20] \\
\hline
\end{tabular}
\caption{Performance of models under Color and Size tasks with two flip directions: WK $\to$ CF and CF $\to$ WK. The key layers are shown for each flip direction.}
\label{tab:flips}
\end{table}

\begin{figure*}[h!]
    \centering
    \includegraphics[width=0.85\textwidth]{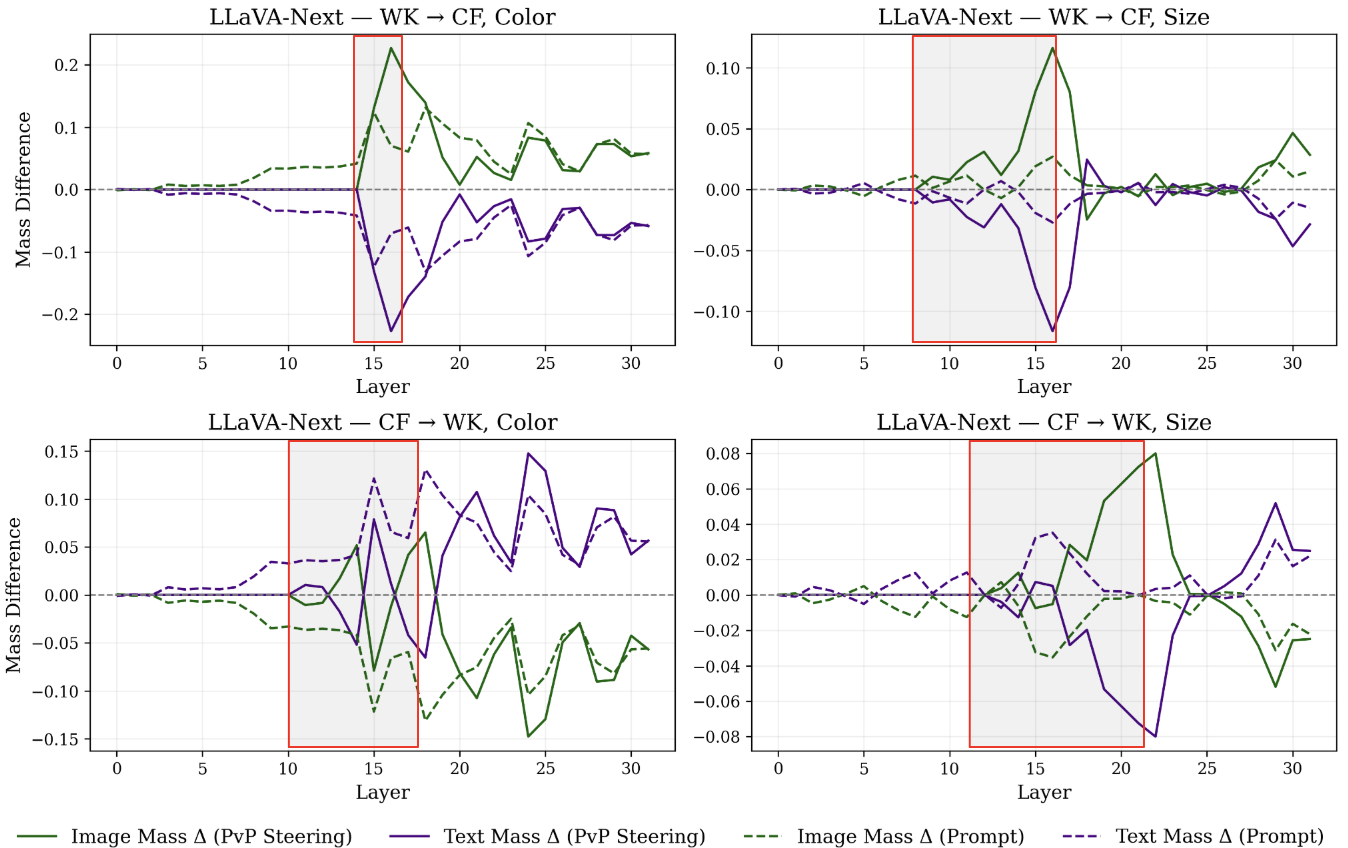}
    \caption{
    Effect of prompt changes and interventions on attention mass across layers for LLaVA-Next in the color and size tasks. Solid lines show changes when applying the steering vector; dashed lines show the effect of modifying the prompt. Green and purple lines represent attention shifts toward image and text tokens, respectively. The red shaded region highlights the layers where the intervention was applied (corresponding to Table \ref{tab:flips}). We see that intervention has a much stronger effect than changing the prompt. 
    }
    \label{fig:llava_attn}
\end{figure*}

In Section~\ref{sec:early_decoding_results}, we show that MLLMs tend to rely on world knowledge in early layers and shift to visual information later, often flipping between the two. This delayed integration of visual input often results in unstable predictions when images conflict with prior knowledge (seen in Table ~\ref{tab:accuracies}). To stabilize predictions and control whether the model attends to the image or draws from prior knowledge, we use Pixel Versus Prior Steering (see Section~\ref{sec:pvp_steering}). Practically, PvP steering offers an interpretable method to causally intervene in model processing, revealing the layers and temporal windows where the balance between vision and world knowledge can be effectively manipulated.

Table~\ref{tab:flips} shows the effectiveness of our steering approach across tasks and models, highlighting both the percentage of successful steering of the model and the key layers at which intervention has the highest impact. We apply PvP steering vectors to the set of inputs that the model originally gets incorrect, meaning without PvP-steering, the model performance on this subset of data is 0\%. Remarkably, we achieve at least a 98\% success rate in flipping model predictions from world knowledge to counterfactual answers in the color task for all models. This demonstrates that MLLMs are not only steerable but highly responsive to targeted interventions, particularly when guided away from strongly encoded world knowledge priors. The size task is more difficult by nature: it requires detecting two objects and reasoning about their size relationship, making it more dependent on deeper, distributed visual processing. This is reflected in lower flip rates and broader intervention windows. 

Across models, we observe that the most effective interventions tend to occur within specific mid-to-late layer ranges, typically requiring sustained influence over multiple layers (Table~\ref{tab:flips}). In general, steering the model from world knowledge to a counterfactual (WK~$\rightarrow$~CF) demands less intervention than reversing that shift (CF~$\rightarrow$~WK), suggesting that overriding memorized priors is easier than restoring them once suppressed by a counterfactual input. 

%Figure~\ref{fig:pca-steering} visualizes the top two principal components of the last hidden states from LLaVA-Next, comparing representations before and after PvP steering. Each point corresponds to a hidden state obtained from either the prompt ``most'' (world knowledge) or ``this'' (pixel-based), with and without PvP. Steering from world knowledge to counterfactual answers (orange line $\rightarrow$ blue line) shifts the representations along a direction closely aligned with the counterfactual-guided prompt. In contrast, steering from pixel-based prompts to world knowledge prior answers (green $\rightarrow$ purple) induces a shift that is less aligned with the original ``most'' prompt direction, indicating greater representational reorientation is required when steering from counterfactual answers to world knowledge. These patterns align with the results in Table \ref{tab:flips}, highlighting how both task complexity and the scope of intervention determine when and where the model’s attention can be effectively redirected.

\subsection{Impact of Prompting and Steering Vectors on Attention Mass}

\begin{figure}[h]
    \centering
    \includegraphics[width=\columnwidth]{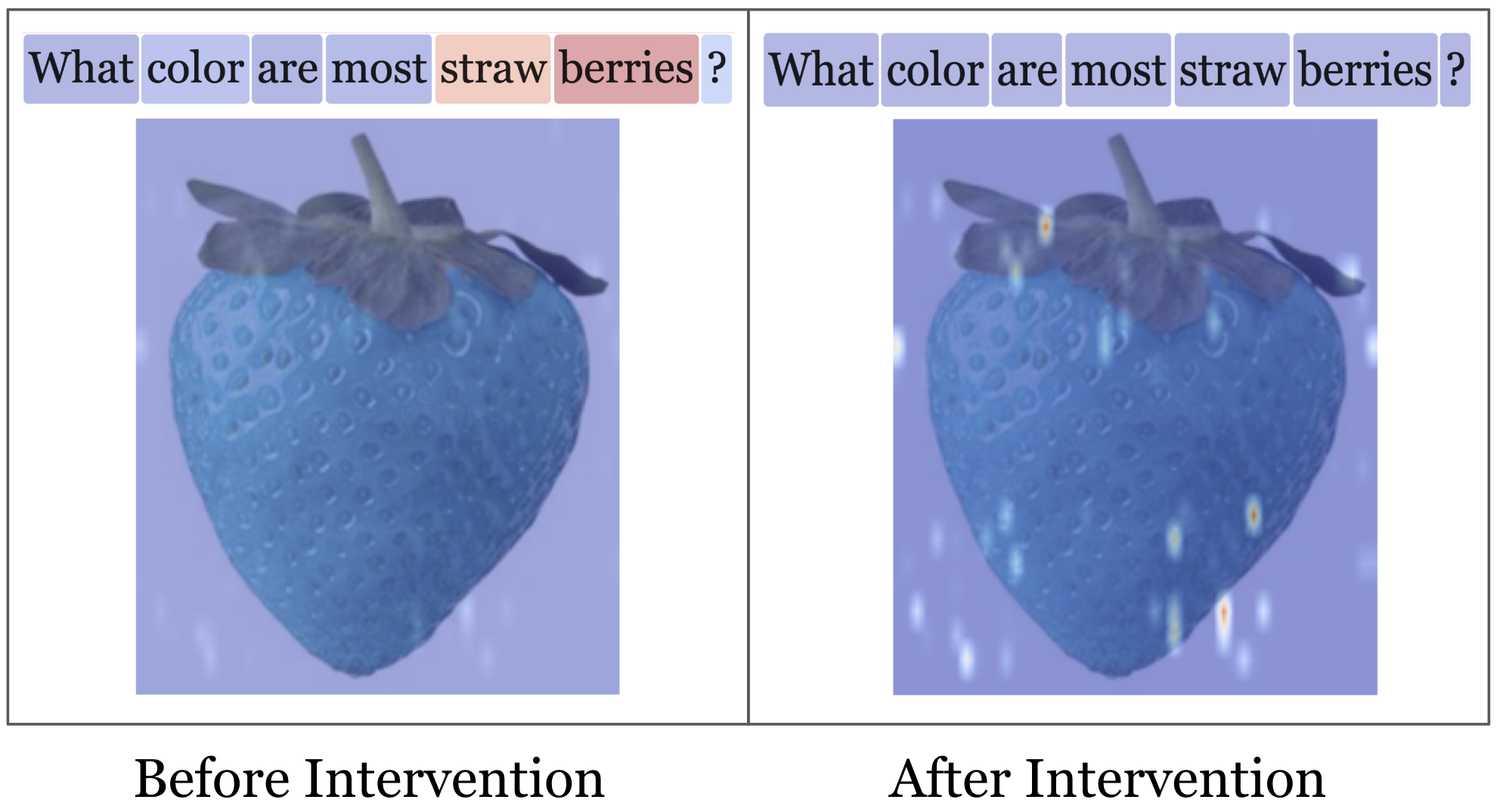}
    \caption{
    Impact of intervention on attention distribution. We observe more attention on text before intervention, and see a shift in attention towards the image and away from the text after intervention. 
    }
    \label{fig:attn_strawberry}
\end{figure}
% \begin{figure}[h!]
%     \centering
%     \includegraphics[width=\columnwidth]{figures/steering_pca.png}
%     \caption{First two principal components of sentence embeddings of LLaVA-Next before and after steering from priors to pixels and from pixels to priors.}
%     \label{fig:pca-steering}
% \end{figure}

% \begin{figure}[h]
%     \centering
%     \includegraphics[width=\columnwidth]{figures/strawberry_attention_map_2.png}
%     \caption{
%     Impact of intervention on attention distribution. We observe more attention on text before intervention, and see a shift in attention towards the image and away from the text after intervention. 
%     }
%     \label{fig:attn_strawberry}
% \end{figure}

While our results demonstrate that PvP steering vectors can reliably shift model outputs, we do not know how this shift is implemented internally. We hypothesize that this shift is implemented in the attention layers, as these components gather information from either the image or text tokens. To study the impact of steering vectors on model predictions, we analyze their impact on the model's attention patterns. We compare two settings: (1) changing the prompt from ``most'' to ``this'' (dashed lines), and (2) applying our PvP steering vector (solid lines) that steers the model toward or away from a counterfactual response. The first setting explores how changing the prompt shifts the model’s attention. Asking about the color of ``most'' strawberries should encourage the model to focus on prior knowledge (red), while asking about ``this'' strawberry directs attention to the specific pixels (blue). The second setting shows how injecting the PvP steering vector guides the model’s internal attention beyond the effect of the prompt.

As shown in Figure~\ref{fig:llava_attn}, changing the prompt from ``most'' to ``this'' yields a modest shift in attention toward image tokens. For example, on the color task, LLaVA-Next increases the attention mass to image tokens by 13\%. In contrast, the PvP intervention vector causes a much stronger shift, increasing attention mass to image tokens to 40\% (see Table~\ref{tab:attn_mass} in Appendix \ref{app:attn_mass} for all models).

Among all models, LLaVA-Next shows the strongest and most consistent shifts in attention when interventions are applied, followed by Qwen2-VL and then Janus-Pro. For the size task, steering vectors must act earlier in the network and across more layers to be effective, reflecting the fact that size requires integrating more visual features than color. In contrast, the color task is more localized and easier to influence with a smaller intervention window. Figure~\ref{fig:llava_attn} illustrates how attention moves across layers in response to both prompt changes and steering task interventions in LLaVA-Next. To illustrate the same effect but with a concrete example, Figure \ref{fig:attn_strawberry}, shows the attention before intervention being heavily focused on text, with ``strawberries'' highlighted in red (highest attention). After intervention, attention shifts to the image, with most red regions being inside the image, rather than over the text. 

These findings show that PvP steering vectors reshape internal attention mechanisms more effectively than prompt changes alone. They offer precise control over how models allocate attention to visual inputs, especially in tasks like size comparison that require broader spatial reasoning. By intervening directly in the model’s representation space, PvP steering enables deeper interpretability and control over MLLM behavior.

%Table~\ref{tab:image_mass_change} quantifies these effects across models and tasks, while Figure~\ref{fig:llava_attn} illustrates how attention moves across layers in response to both prompt changes and steering task interventions in LLaVA-Next.
%To illustrate the impact of the intervention compared to the direct prompt, we visualize attention patterns overlaid on an image-text input example. Figure \ref{fig:attn_strawberry}, shows the attention before intervention being heavily focused on text, with ``strawberries'' highlighted in red (highest attention). After intervention, attention shifts to the image, with most red regions being inside the image, rather than over the text. 

\section{Conclusion}
In this work, we investigate how multimodal large language models (MLLMs) reconcile memorized world knowledge and visual input. Understanding this balance is essential for building reliable models that can correctly choose between conflicting sources of information. To study this, we introduce \textit{Visual CounterFact}, a dataset that constructs realistic visual counterfactuals targeting familiar attributes like object color and size. These examples violate learned priors while preserving visual plausibility, enabling precise comparisons between perception and memory. Using this dataset, we find that MLLMs often default to perception, even when prompted to retrieve general knowledge. In these cases, performance on knowledge-based prompts drops significantly, suggesting that models are overly influenced by visual inputs, even when the question targets memorized facts.
Through studying the forward-pass, we observe that model predictions initially reflect stored priors, then transition to visually grounded answers in mid-to-late layers. This transition is often unstable, with models flipping between the two sources of information across layers.
To control this behavior, we introduce \textit{Pixels Versus Priors} steering vectors, which allow us to edit model behavior toward preferring either world knowledge priors or visual input. These activation-level interventions produce significant attention shifts towards or away from the image, depending on our steering vector direction. Our findings offer a new framework for interpreting and controlling MLLMs, advancing our ability to understand and control the interaction between memory and perception in multimodal models.

\section{Limitations}
Our framework focuses on three state-of-the-art models: LLaVA-Next, Qwen2-VL, and Janus-Pro, which, while diverse, do not represent the full spectrum of multimodal architectures, such as monolithic MLLMs. However, this level of focus is consistent with standard practice in interpretability research, where analyses typically target one or two models to enable detailed, mechanism-level insights across both LLMs and MLLMs \cite{rome, dai2021knowledge, luo2024task, hojel2024finding}. Despite architectural differences, our findings consistently generalize across the models studied, supporting the robustness of our approach. In future work, we plan to expand our analysis to a broader range of models to explore how architectural design impacts reliance on perception versus prior knowledge.

Additionally, through our analysis we find that steering models from visual perception back to world knowledge is more difficult than the reverse, suggesting an asymmetry in how MLLMs prioritize in-context versus memorized information. Understanding this distinction further remains an open direction for future work.

\bibliography{bib}

\appendix

\section{Creating Visual CounterFact}\label{app:dataset_construction}
\begin{figure*}[h]
    \centering
    \includegraphics[width=\textwidth]{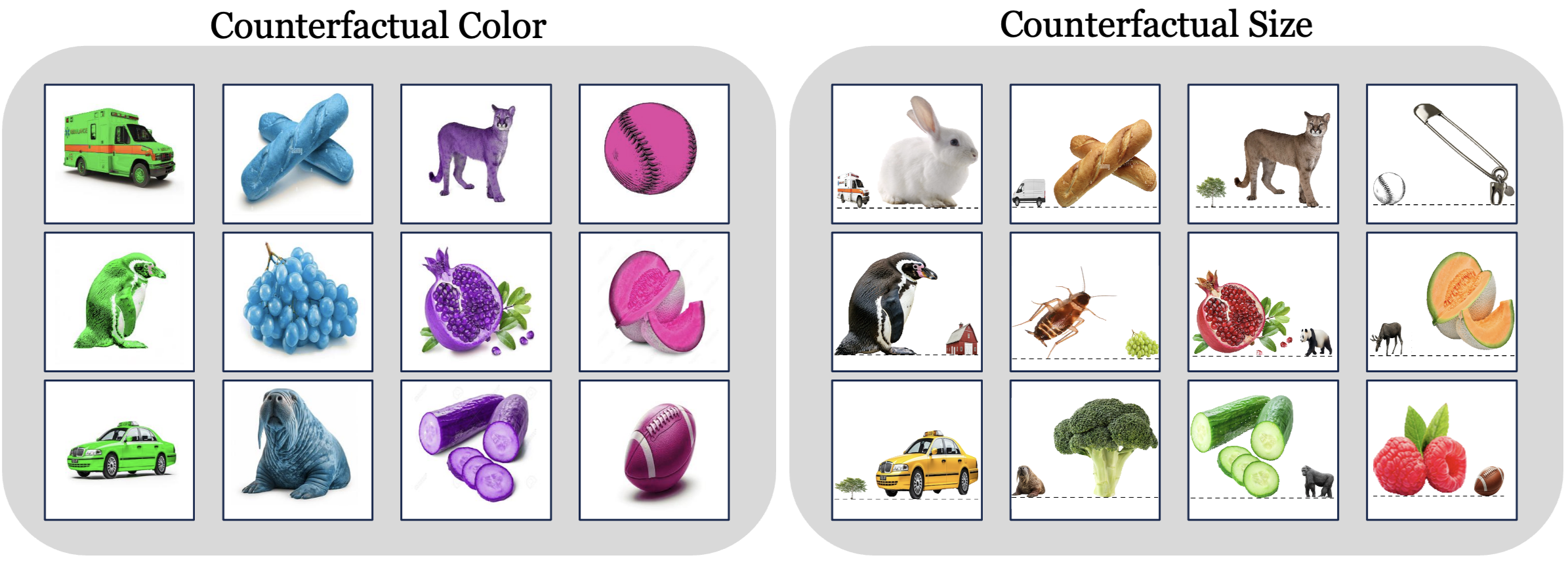}
    \caption{Examples from our dataset. (Left) images created using color relations, 
    (right) images created using size relations.}
    \label{fig:more_examples}
\end{figure*}

\begin{figure}[h]
    \centering
    \includegraphics[width=0.85\columnwidth]{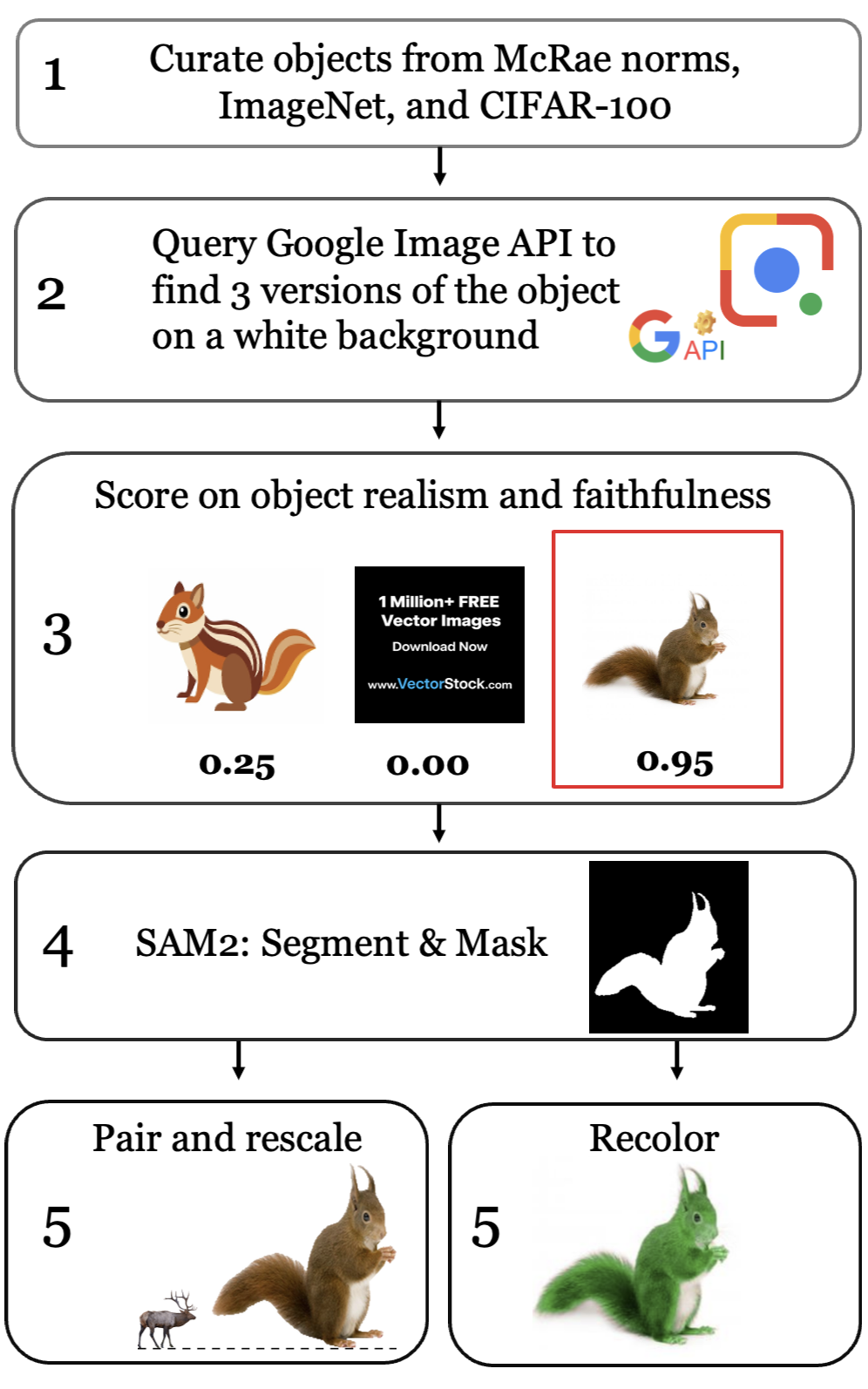}
    \caption{Construction pipeline for the Visual Counterfactual Dataset. We identify typical traits using semantic knowledge sources, retrieve realistic visual exemplars, and apply transformations to create perceptually plausible counterfactuals that conflict with language priors.}
    \label{fig:dataset_pipeline}
\end{figure}

\textbf{Step 1: Sourcing Objects.} We begin by identifying a set of real-world \textit{objects} (e.g., \textit{strawberry}, \textit{squirrel}, \textit{cherry}) from the McRae feature norms dataset \cite{mcrae2005semantic}, ImageNet \cite{imagenet}, and CIFAR-100 \cite{cifar100}. The McRae dataset tasks 100 human participants with listing common attributes for an object. If at least 30\% of participants respond with a specific color as a key attribute of that object, we include it in our dataset. After filtering with this 30\% threshold, we obtain 190 objects. To increase the number of objects, we use the GPT-4o to elicit typical colors for ImageNet 1000 and CIFAR-100 classes and discard any categories lacking strong color ground-truth priors (e.g., clothing items). After both procedures, we obtain \textbf{622} unique objects. 

\textbf{Step 2: Sampling Real-World Images.} We use the Google Images API to retrieve three candidate images of each object, specifying that they must be on a white background (a \texttt{<world knowledge color>} \texttt{<subject>} on a white background). Since prior work has shown that background information can heavily bias object detection models, we extract objects against a white background to ensure that model decisions are not influenced by spurious background features \cite{noise_signal, spuriosityrankingssortingdata, spurious_correlation_mllm, perception_clip}. Sampled images are then evaluated using GPT-4o, which scores them based on object correctness, color accuracy, presence of a white background, and overall realism. Specifically, we prompt GPT-4o with the following questions:

\begin{quote}
\begin{small}
\begin{enumerate}
    \item Is this an image of a \texttt{<color>} \texttt{<object>}? (yes/no)
    \item Is this image on a white background? (yes/no)
    \item Is this image an illustration or a realistic image? (illustration/realistic)
    \item On a scale from 1 to 10, how realistic is this \texttt{<object>}? (numerical score)
\end{enumerate}
\end{small}
\end{quote}

We retain the highest-scoring image of the three to ensure visual fidelity with our inclusion criteria. For each yes/no question, the image receives a score of 0 for ``no'' or a score of 10 for ``yes''. For images scoring 0 overall, we repeat the querying process but remove the world knowledge color (query: a \texttt{<subject>} on a white background), as most of the images resulting in a score of 0 are multi-colored (e.g., zebra, bee). After the second round of querying, we drop any remaining images with score 0, resulting in \textbf{575} unique objects. 

\textbf{Step 3: Constructing Object-Size and Object-Color Counterfactuals.}

\textbf{(1) Color:} 
To generate color counterfactuals, we prompt the LLaVA-Next model (Mistral-7B backbone) with ``What color is a \texttt{<object>}?'' and randomly sample from the five least likely color predictions (using common colors such as red, blue, green, pink, orange, etc.). This ensures that counterfactuals challenge the model’s linguistic priors, encouraging reliance on visual input rather than memorized associations. We constrain counterfactual colors to be perceptually distinct from the original color (e.g., avoiding red/orange or gray/black swaps). 
\textbf{(2) Size:} 
For the size task, we use the same set of objects from the color task and estimate their typical real-world dimensions using GPT-4o. The model provides height and width in inches, which we multiply to compute a total size metric. We then identify object pairs that differ in size by at least a factor of 10. For each object, we create two counterfactual images. Given three objects that satisfy \texttt{object$_1$ < object$_2$ < object$_3$}, where ``<'' denotes increasing real-world size, we generate two counterfactual images containing \texttt{object$_2$}. Namely,  \texttt{(object$_1$, bigger\_than, object$_2$)} and \texttt{(object$_2$, bigger\_than, object$_3$)}. For example, if a squirrel is typically bigger than a cherry and smaller than an alligator, we create the counterfactual images \texttt{(cherry, bigger\_than, squirrel)} and \texttt{(squirrel, bigger\_than, alligator)}. This creates twice the number of samples since we construct two size relations for one object. After manual filtering of sizes that GPT-4o reported incorrectly, we are left with 877 unique size-object relations. 

\textbf{Step 4: Creating Counterfactual Images.} Given a retrieved object on a white background, the first step in creating counterfactual images is to use SAM2 \cite{sam2} to obtain a segmentation mask. After we obtain segmentation masks, we use two separate pipelines to create color and size counterfactuals, respectively.

\textbf{(1) Color:} Given a color-object relation with the relation ``\texttt{[object]} is \texttt{[color]}'' and its counterfactual relation ``\texttt{[object]} is \texttt{[counterfactual color]}'', we apply a segmentation mask to isolate the object and modify only hue values in the HSV color space in order to change the color of the object to the counterfactual color while preserving the original saturation and brightness. This produces realistic and semantically surprising transformations (e.g., turning a red strawberry blue) while maintaining texture and shading. For objects with minimal hue (e.g., gray or black), we apply a set of hand-written remapping rules. 

\textbf{(2) Size:} 
Given two objects with the world knowledge relation ``\texttt{[object 1]} is larger than \texttt{[object 2]}'' we create the counterfactual image reflecting ``\texttt{[object 2]} is larger than \texttt{[object 1]}'' by combining the segmentation masks of object$_1$ and object$_2$. For the world knowledge image, we resize the masks so that object$_1$ appears significantly larger than object$_2$, and in the counterfactual image, we resize the masks so that object$_2$ appears significantly larger than object$_1$. Specifically, 250$\times$250 pixels versus 80$\times$80 pixels. This size difference visually reflects the intended relation.

To make the size comparison clear and avoid depth ambiguity, we place both objects on the same horizontal baseline and add a black dashed reference line that both objects touch. This helps ensure that differences in perceived size are interpreted as scale changes rather than perspective shifts. 

Visual CounterFact consists of 575 original (world knowledge) object images, 575 color counterfactual images, and 877 size original and 877 counterfactual images, totaling 2,904 unique images. Figure~\ref{fig:counterfactuals} shows examples from each split of the dataset. These transformations yield a dataset that explicitly conflicts with world knowledge priors of an object's color and size while preserving perceptual plausibility, enabling targeted evaluation of visual reasoning models under counterfactual conditions. In Appendix Section~\ref{app:dataset_examples}, we provide additional examples of images as well as dataset statistics on the kinds of objects we include in Visual Counterfact.

\section{Dataset Statistics and Examples}
\label{app:dataset_examples}

\begin{table}[h]
\centering
\begin{tabular}{l r}
\toprule
\textbf{Category} & \textbf{Count} \\
\midrule
Animals & 218 \\
Household Items and Furniture & 59 \\
Fruits and Vegetables & 58 \\
Vehicles and Transportation & 42 \\
Electronics and Appliances & 29 \\
Tools and Hardware & 22 \\
Food and Drink (non-produce) & 21 \\
Buildings and Structures & 21 \\
Plants and Trees & 19 \\
Musical Instruments & 15 \\
Clothing and Accessories & 15 \\
Weapons and Military Items & 13 \\
Medical and Hygiene Items & 11 \\
Toys and Recreational Items & 11 \\
Natural Objects (non-living) & 10 \\
Office Supplies & 6 \\
Miscellaneous & 5 \\
\bottomrule
\end{tabular}
\caption{Distribution of object categories in the dataset.}
\label{tab:object_categories}
\end{table}

Figure~\ref{fig:counterfactuals} illustrates representative examples from the Visual CounterFact dataset, including counterfactual edits based on object color (left) and relative size (right). Each example maintains visual realism while introducing semantically meaningful contradictions to typical object properties.

Table~\ref{tab:object_categories} summarizes the distribution of object categories in the dataset. The majority of counterfactuals involve animals, followed by a diverse set of objects spanning furniture, produce, vehicles, tools, and more. This broad coverage ensures the dataset tests model reliance on both visual input and memorized associations across varied semantic domains.

\section{Attention Mass Details}\label{app:attn_mass}

\begin{figure*}[h!]
    \centering
    \includegraphics[width=0.98\textwidth]{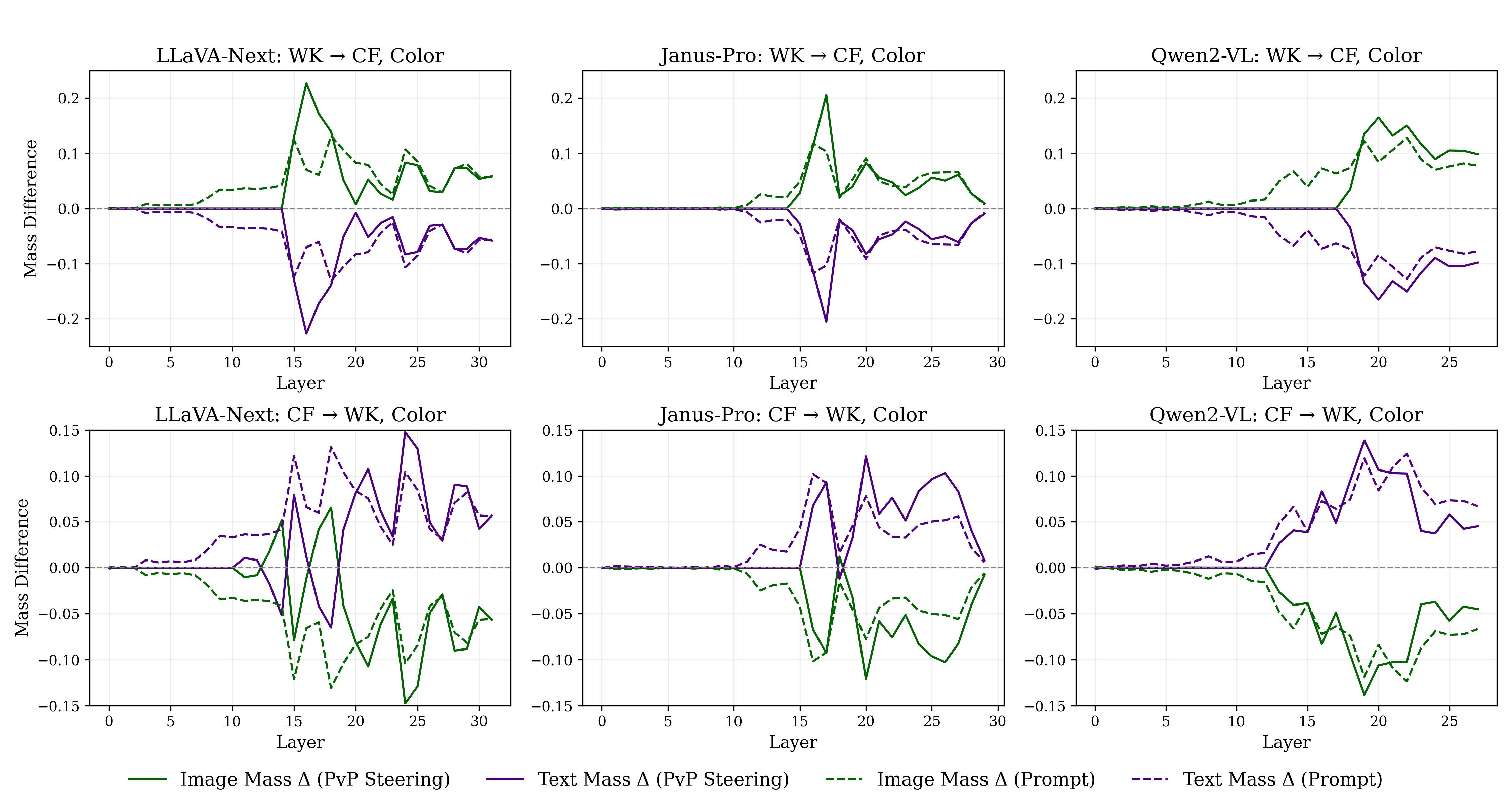}
    \caption{Attention mass difference across layers for all models on the color task. Solid lines show changes from PvP steering vectors; dashed lines show prompt-only changes. Green represents attention to image tokens, purple to text tokens. Each subplot shows one model and intervention direction (WK~$\rightarrow$~CF or CF~$\rightarrow$~WK).}
    \label{fig:attn_mass_color_all}
\end{figure*}

\begin{figure*}[h!]
    \centering
    \includegraphics[width=0.98\textwidth]{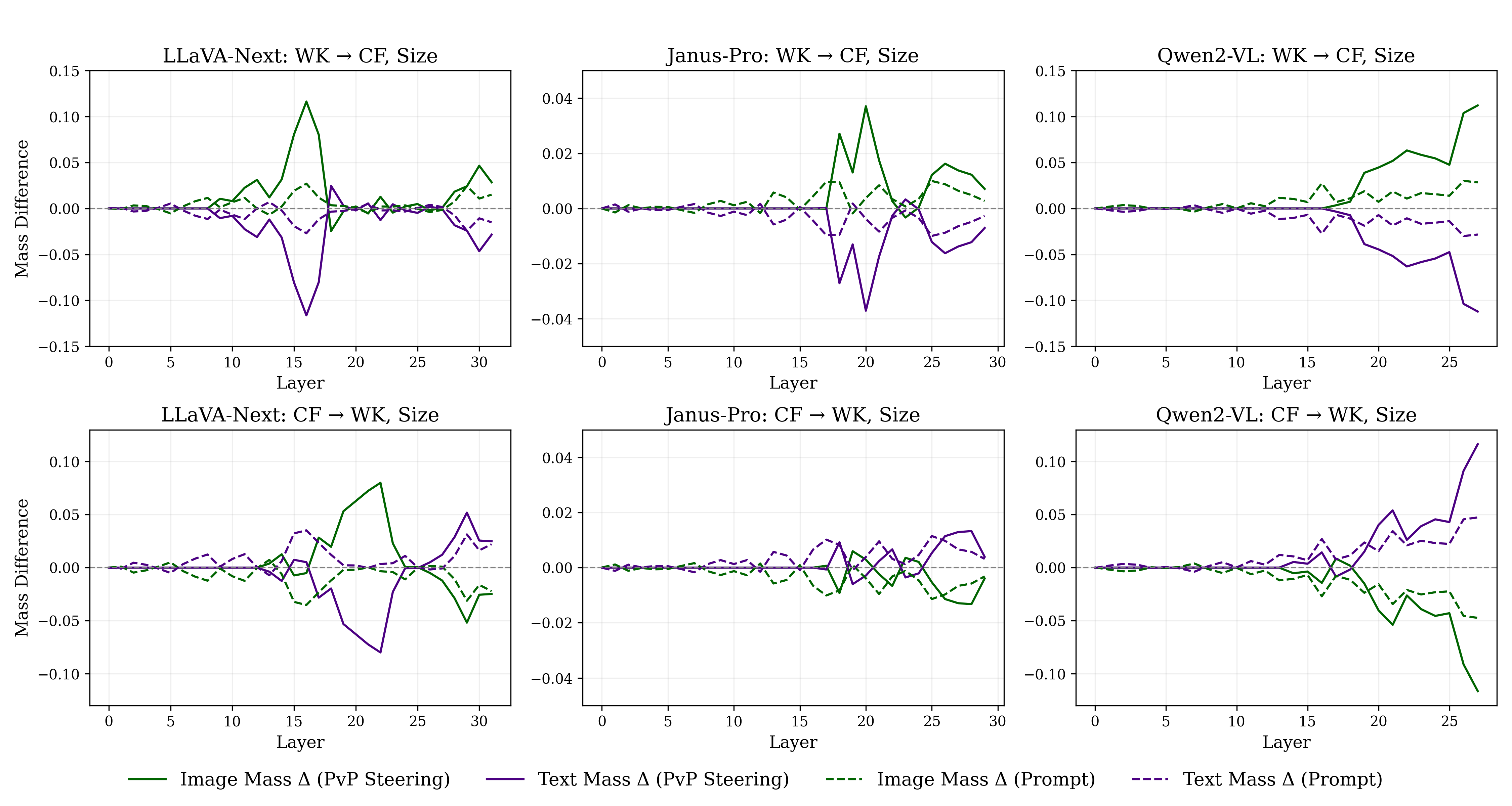}
    \caption{Attention mass difference across layers for all models on the size task. Solid lines show changes from PvP steering vectors; dashed lines show prompt-only changes.  Each subplot shows one model and intervention direction (WK~$\rightarrow$~CF or CF~$\rightarrow$~WK).}
    \label{fig:attn_mass_size_all}
\end{figure*}

To better understand how steering vectors and prompt changes affect the internal attention mechanisms of MLLMs, we visualize the change in attention mass over layers for each model and task. Figures~\ref{fig:attn_mass_color_all} and~\ref{fig:attn_mass_size_all} show these effects across color and size tasks, respectively. Each subplot compares attention mass difference toward image tokens (green) and text tokens (purple), with solid lines indicating PvP steering interventions and dashed lines indicating prompt-only changes. 

We find that across all models and both tasks, interventions consistently produce stronger shifts in attention mass compared to prompt changes alone. In the color task (Figure~\ref{fig:attn_mass_color_all}), steering from WK to CF reliably increases image attention, while the reverse direction decreases it (as expected). The effect is particularly pronounced in LLaVA-Next, with peaks around the intervention window. The size task (Figure~\ref{fig:attn_mass_size_all}) shows a similar but more muted pattern, consistent with the task's higher visual complexity. These trends reinforce that PvP steering vectors exert precise, causal control over how attention is allocated between vision and language streams.

\begin{figure}[h]
    \centering
    \includegraphics[width=\columnwidth]{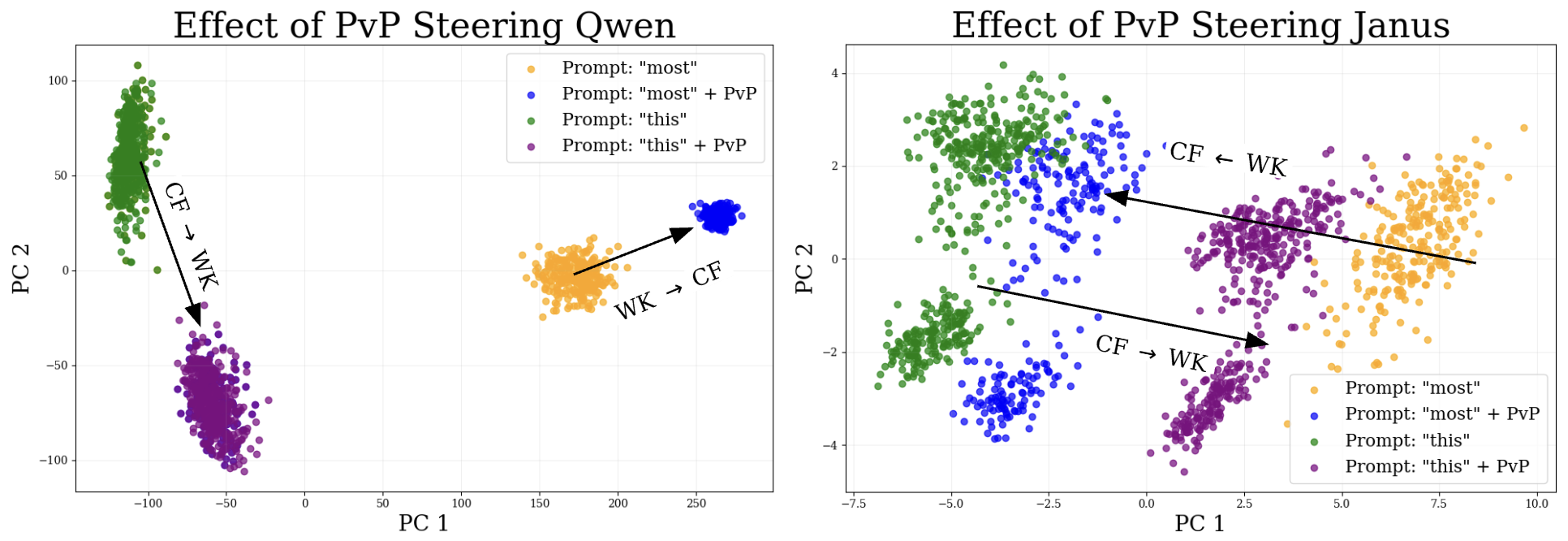}
    \caption{First two principal components of sentence embeddings of Qwen2.5-VL and Janus-Pro before and after steering from priors to pixels and from pixels to priors.}
    \label{fig:qwen_janus_steering}
\end{figure}

\begin{table}[h]
\centering
\small  % Reduces font size slightly
\begin{tabular}{l l r r}
\toprule
\textbf{LLaVA-Next} & & \textbf{Intervention} & \textbf{Prompt} \\
\midrule
Color & WK $\rightarrow$ CF & $40.0\%$ & $13.1\%$ \\
Color & CF $\rightarrow$ WK & $-15.6\%$ & $-13.1\%$ \\
Size  & WK $\rightarrow$ CF & $10.9\%$ & $3.2\%$ \\
Size  & CF $\rightarrow$ WK & $-7.6\%$ & $-3.2\%$ \\
\midrule
\textbf{Qwen2-VL} & & & \\
\midrule
Color & WK $\rightarrow$ CF & $21.8\%$ & $12.8\%$ \\
Color & CF $\rightarrow$ WK & $-25.2\%$ & $-12.4\%$ \\
Size  & WK $\rightarrow$ CF & $14.1\%$ & $4.7\%$ \\
Size  & CF $\rightarrow$ WK & $-10.8\%$ & $-4.7\%$ \\
\midrule
\textbf{Janus-Pro} & & & \\
\midrule
Color & WK $\rightarrow$ CF & $19.5\%$ & $11.7\%$ \\
Color & CF $\rightarrow$ WK & $-11.0\%$ & $-10.2\%$ \\
Size  & WK $\rightarrow$ CF & $2.4\%$ & $1.1\%$ \\
Size  & CF $\rightarrow$ WK & $-1.1\%$ & $-1.1\%$ \\
\bottomrule
\end{tabular}
\caption{Max change in image attention mass ($\Delta$) under intervention and prompt changes for each model and task.}\label{tab:attn_mass}
\end{table}

Table~\ref{tab:attn_mass} reports the maximum change in attention mass directed toward image tokens across models and tasks, comparing the effects of prompt changes (``most'' to ``this'') and PvP steering interventions. As described in Section~\ref{sec:pvp_steering}, we measure the peak increase or decrease in image attention during inference.

Across all models, PvP steering consistently produces larger attention shifts than prompt modifications. This effect is most pronounced in the color task, where interventions increase image attention by up to 40\% in LLaVA-Next, compared to 13\% from prompting. As visualized in Figure~\ref{fig:llava_attn}, prompting leads to moderate reallocation of attention, while steering vectors induce strong and targeted redistribution.

We also observe an asymmetry between WK~$\rightarrow$~CF and CF~$\rightarrow$~WK directions: steering toward perception (WK~$\rightarrow$~CF) is generally more effective than restoring attention to prior-based information (CF~$\rightarrow$~WK). This aligns with accuracy results in Section~\ref{sec:steering_results}, where interventions that shift models away from priors are more successful than those that attempt to recover them.

These findings provide further evidence that PvP steering offers fine-grained control over internal attention dynamics in MLLMs, outperforming prompt-based techniques in both strength and specificity.

\section{Early Decoding}

\begin{figure*}[t]
    \centering
    \begin{minipage}[b]{0.48\textwidth}
        \centering
        \includegraphics[width=\linewidth]{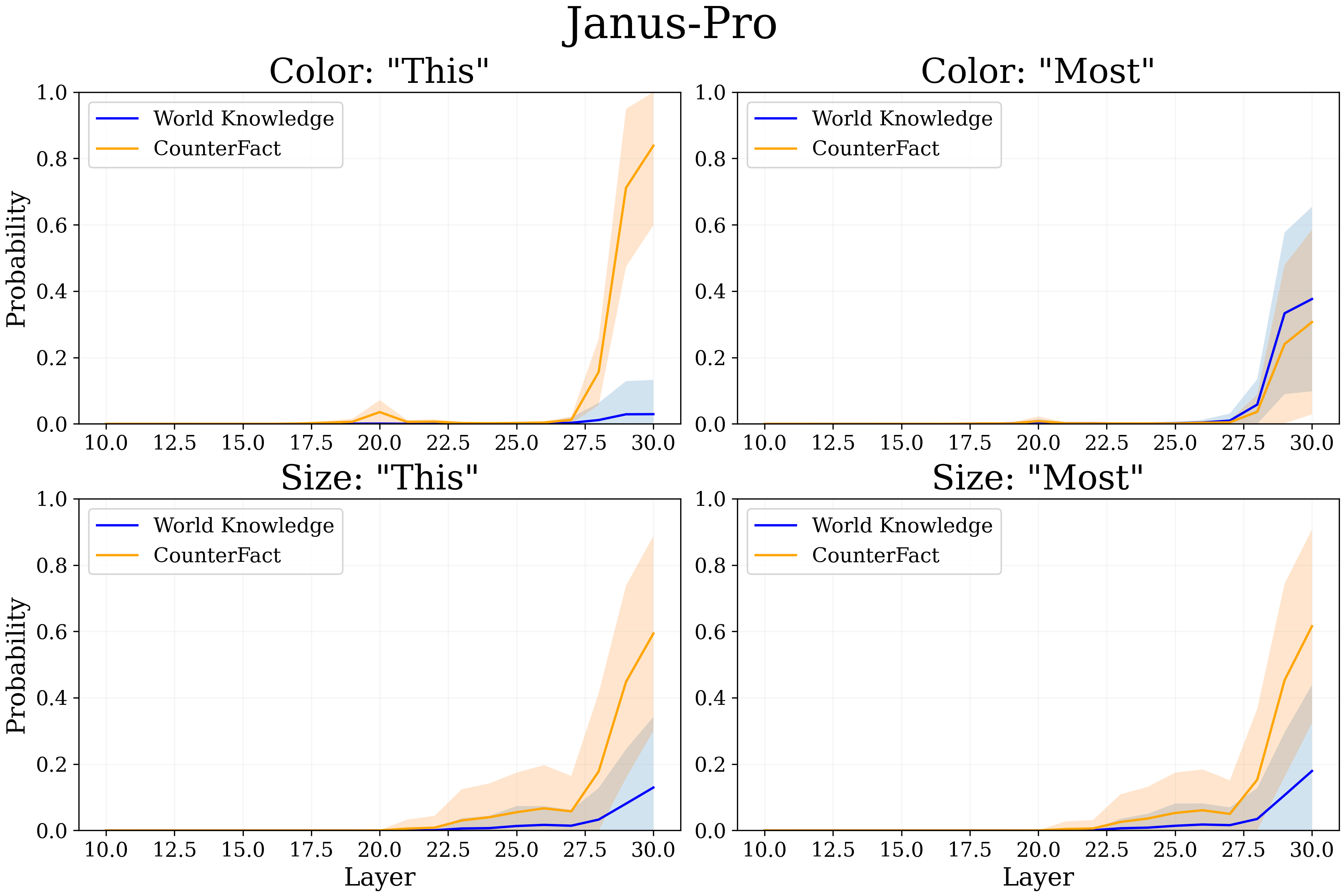}
        \caption*{Janus-Pro}
    \end{minipage}
    \hfill
    \begin{minipage}[b]{0.48\textwidth}
        \centering
        \includegraphics[width=\linewidth]{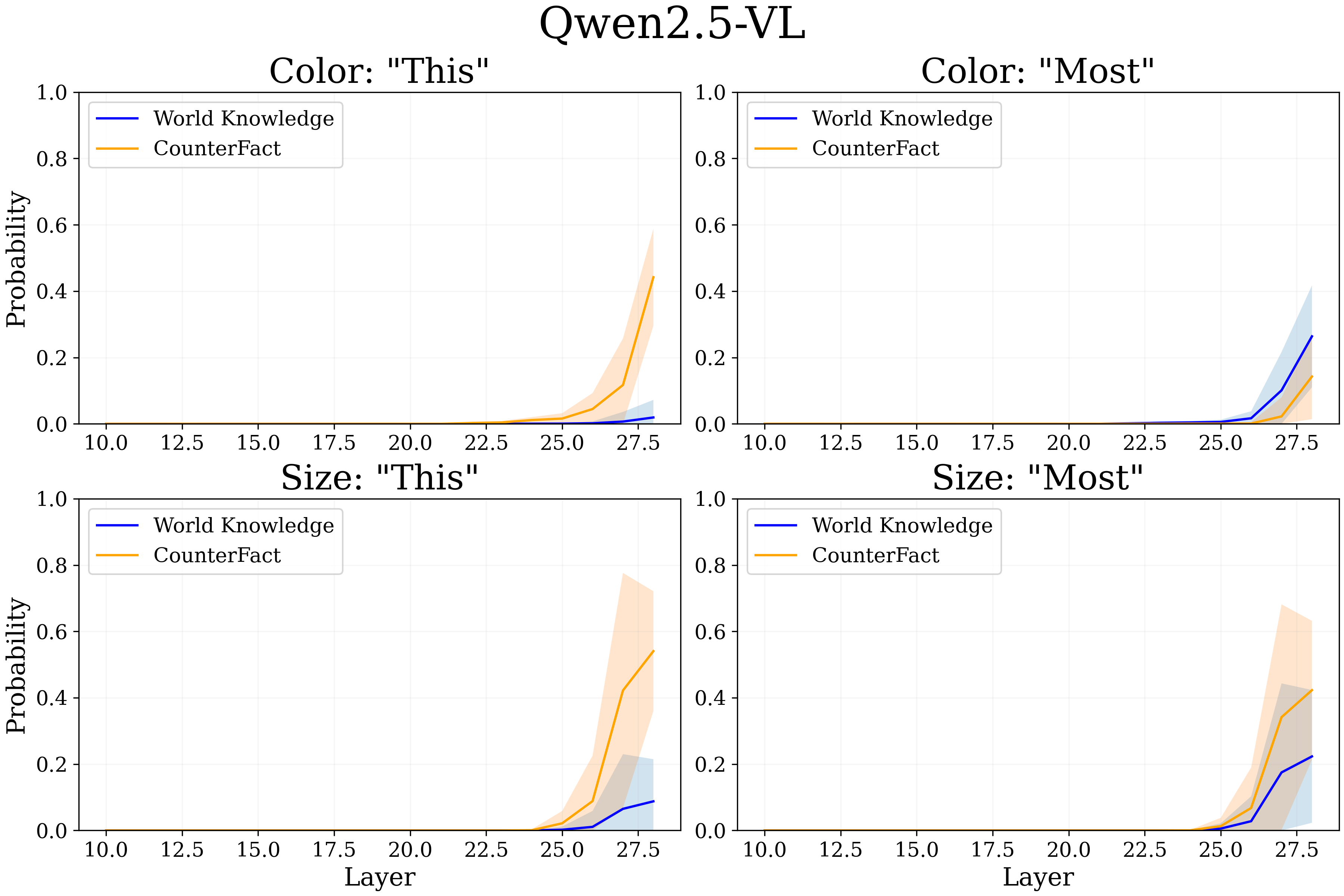}
        \caption*{Qwen2.5-VL}
    \end{minipage}
    \caption{Early decoding results on Qwen2.5-VL and Janus-Pro show a conflict between answering ``world knowledge'' using in-weight memorization or answering ``counterfact'' using visual perception.}
    \label{fig:early_decoding_appendix}
\end{figure*}

Figure~\ref{fig:early_decoding_appendix} shows early decoding traces for Qwen2.5-VL and Janus-Pro, extending our main-layer analysis from Figure~\ref{fig:early_decoding}. Consistent with the behavior observed in LLaVA-Next, both models initially assign high probability to the world knowledge answer when prompted with a ``most'' question and shown a counterfactual image. However, as the forward pass progresses, the probability of the counterfactual answer rises and ultimately dominates by the final layer.

For both models, this flipping behavior illustrates the delayed integration of visual information, often leading the model to override its prior with perceptual evidence late in the forward pass. When prompted with ``this'' questions, both models quickly favor the counterfactual answer and rarely flip to world knowledge. These early decoding results across all three models reinforce our central finding: MLLMs are highly sensitive to visual input and tend to prioritize perception over memorized priors when the two conflict, particularly in later layers of processing.

%\section{PCA Results}

%Figure~\ref{fig:qwen_janus_steering} shows the principal component analysis (PCA) results for QwenVL and Janus-Pro, complementing the LLaVA-Next results presented in Figure~\ref{fig:pca-steering}. For each model, we project the final hidden state of the last text token into the top two PCA dimensions and visualize the effect of PvP steering vectors. These findings support the asymmetry observed across models and tasks in Table~\ref{tab:flips}. They show that the model’s internal representation space is more easily reoriented away from memorized priors than toward them, further emphasizing the dominance of perception in the presence of conflicting visual input.

\end{document}